\documentclass[10pt,twocolumn,letterpaper]{article}

\usepackage[pagenumbers]{wacv} 

\usepackage[accsupp]{axessibility} 
\usepackage{amsmath,amsfonts,bm}
\usepackage{booktabs}
\usepackage{multirow}
\usepackage{graphicx}

\usepackage[normalem]{ulem}

\newcommand\mypara[1]{\vspace{1mm}\emph{#1}}

\definecolor{citecolor}{rgb}{0,0.08,0.45}
\definecolor{linkcolor}{RGB}{187,18,26}

\definecolor{graftstereo}{RGB}{238, 75, 63}
\definecolor{raftstereo}{RGB}{255, 160, 31}
\definecolor{egdepth}{RGB}{91, 158, 204}
\definecolor{sdgdepth}{RGB}{111, 192, 112}

\def\eqref#1{equation~\ref{#1}}

\def\1{\bm{1}}

\def\vtheta{{\bm{\theta}}}

\def\vb{{\bm{b}}}

\def\mR{{\bm{R}}}

\DeclareMathAlphabet{\mathsfit}{\encodingdefault}{\sfdefault}{m}{sl}
\SetMathAlphabet{\mathsfit}{bold}{\encodingdefault}{\sfdefault}{bx}{n}

\definecolor{wacvblue}{rgb}{0.21,0.49,0.74}
\usepackage[pagebackref,breaklinks,colorlinks,allcolors=wacvblue]{hyperref}

\def\wacvPaperID{3359} 
\def\confName{WACV}
\def\confYear{2026}

\title{Splannequin: Freezing Monocular Mannequin-Challenge Footage with Dual-Detection Splatting}

\author{
Hao-Jen Chien$^{1}$
\quad
Yi-Chuan Huang$^{1}$
\quad
Chung-Ho Wu$^{1}$
\quad
Wei-Lun Chao$^{2}$
\quad
Yu-Lun Liu$^{1}$\vspace{0.5em}
\\
\centerline{$^1$National Yang Ming Chiao Tung University \quad $^2$The Ohio State University}
}

\begin{document}
\twocolumn[{%
\renewcommand\twocolumn[1][]{#1}%
\maketitle
\begin{center}
\centering
\captionsetup{type=figure}
\vspace{-8mm}
\resizebox{0.95\textwidth}{!} 
{
\includegraphics[width=0.9\textwidth]{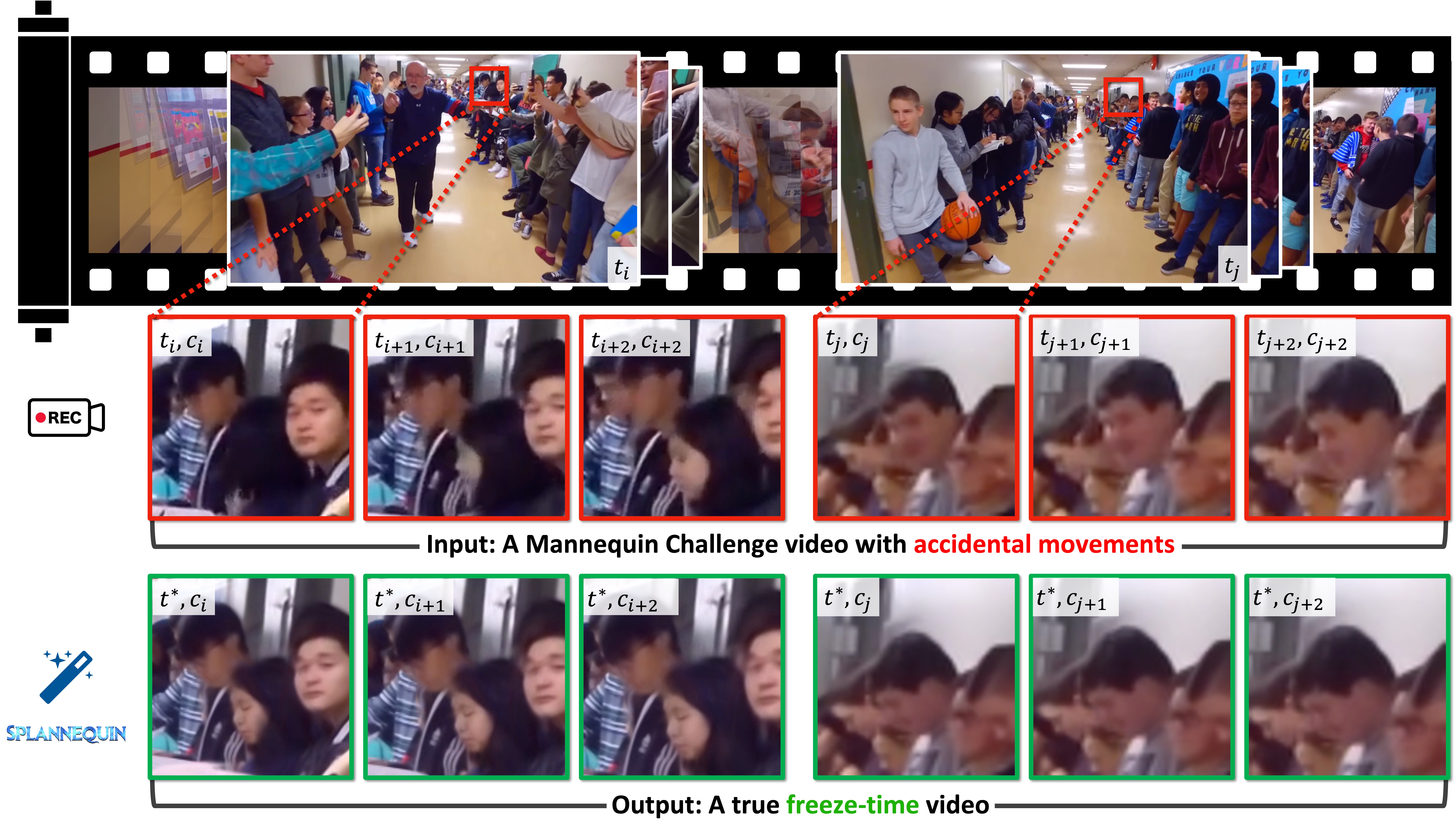}
}
\vspace{-4mm}
\caption{
\textbf{Splannequin converts imperfect Mannequin-Challenge footage into a true freeze-time video.} \emph{(Top)} A monocular \emph{Mannequin-Challenge} video is intended to resemble large-scale frozen frames, yet real-world recordings inevitably contain slight body movements. The \textcolor{red}{\textbf{red}} crops across successive frames with the corresponding camera pose ($c_i$) and timestamp ($t_i$) highlight these noticeable movements. \emph{(Bottom)} After our processing, every crop (\textcolor{Green}{\textbf{green}} boxes) of successive frames remains static. \textbf{Splannequin} analyzes the entire video and resynthesizes a temporally consistent sequence of the same camera poses at $t^\star$ while preserving overall visual fidelity.}
\label{fig:teaser}
\end{center}
}]

\maketitle


\begin{abstract}
Synthesizing high-fidelity frozen 3D scenes from monocular Mannequin-Challenge (MC) videos is a unique problem distinct from standard dynamic scene reconstruction. Instead of focusing on modeling motion, our goal is to create a frozen scene while strategically preserving subtle dynamics to enable user-controlled instant selection. To achieve this, we introduce a novel application of dynamic Gaussian splatting: the scene is modeled dynamically, which retains nearby temporal variation, and a static scene is rendered by fixing the model’s time parameter. However, under this usage, monocular capture with sparse temporal supervision introduces artifacts like ghosting and blur for Gaussians that become unobserved or occluded at weakly supervised timestamps. We propose Splannequin, an architecture-agnostic regularization that detects two states of Gaussian primitives, hidden and defective, and applies temporal anchoring. Under predominantly forward camera motion, hidden states are anchored to their recent well-observed past states, while defective states are anchored to future states with stronger supervision. Our method integrates into existing dynamic Gaussian pipelines via simple loss terms, requires no architectural changes, and adds zero inference overhead. This results in markedly improved visual quality, enabling high-fidelity, user-selectable frozen-time renderings, validated by a 96\% user preference.
Project page: \url{https://chien90190.github.io/splannequin/}

 \end{abstract}
    
\vspace{-1mm}
\section{Introduction}
\label{sec:intro}
\vspace{-1mm}

\textbf{Freeze-time} videos, also known as \emph{Mannequin Challenge} clips, allow cameras to move freely through scenes while subjects appear perfectly frozen mid-action. This can: (1) generate pseudo-ground-truth for training dynamic scene models with cleaner signals~\cite{li2019learning}; and (2) enable artistic control for creators in visual effects (VFX) production to select precise frozen moments. As shown in Figure~\ref{fig:teaser}, we synthesize freeze-time sequences from casual monocular recordings with slight, unintended motion.

Traditional production with multi-camera bullet-time rigs~\cite{sudmann2016bullet} is costly and complex, requiring heavy post-production with single shots costing over \$750k. While the single-camera Mannequin-Challenge is a low-cost alternative, its monocular 3D reconstruction is prone to artifacts from minor subject motion and sparse observational data.

\begin{figure}
    \centering
    \includegraphics[width=0.97\columnwidth]{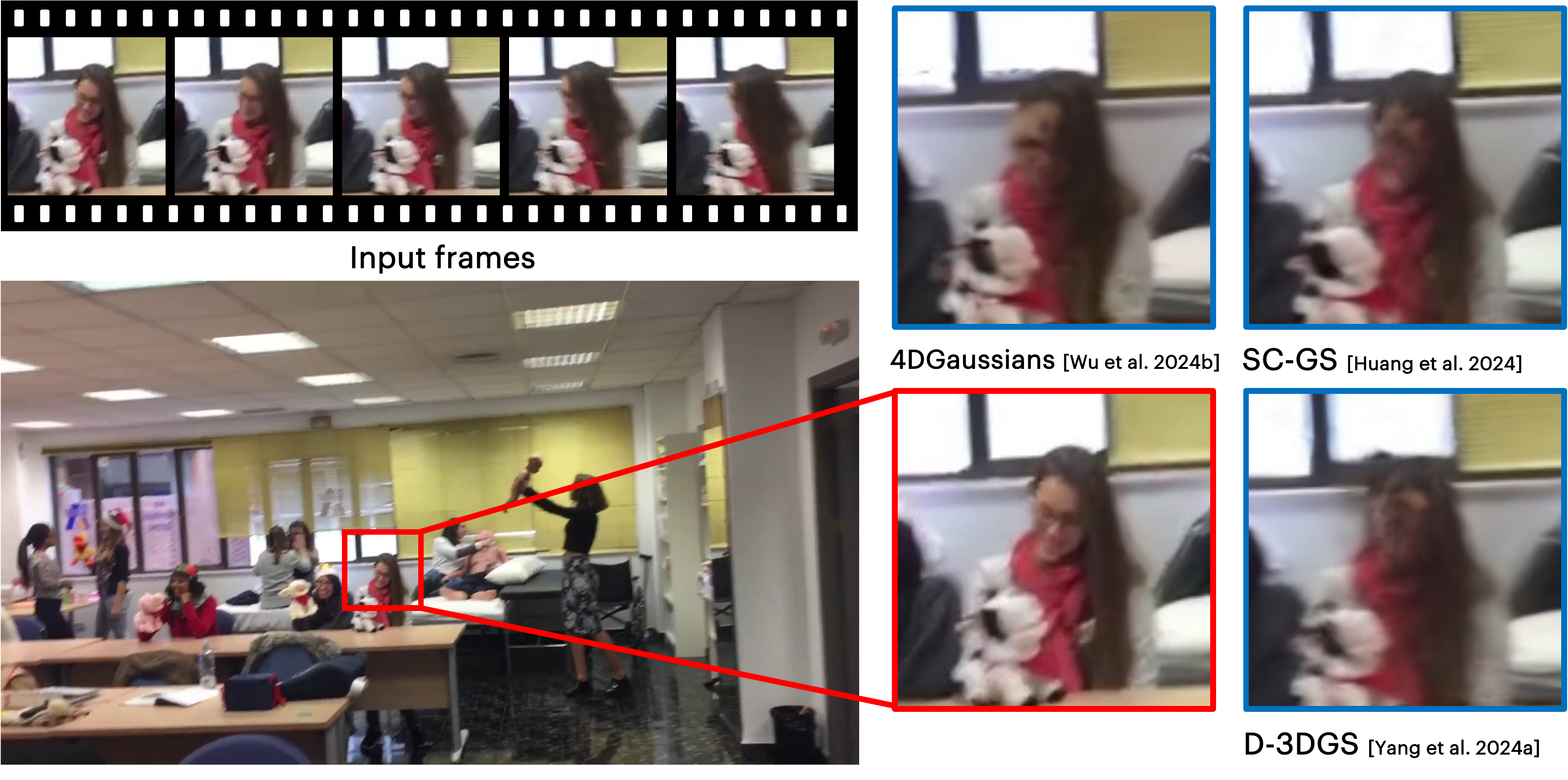}
    \vspace{-3mm}
    \caption{\textbf{Existing Gaussian splatting frameworks cannot produce plausible results from casually captured Mannequin-Challenge videos.} \emph{(Top)} A short clip of the hand-held input sequence exhibits unintentional subject motion. \emph{(Right)} State-of-the-art methods containing SC-GS~\cite{huang2024sc}, 4DGaussians~\cite{wu20244d}, and D-3DGS~\cite{yang2024deformable}. They all leave noticeable blur and double contours around the woman’s face (\textcolor{Blue}{\textbf{blue}} frames). \emph{(Left)} Our Splannequin reconstruction (\textcolor{Red}{\textbf{red}} frame) is crisp and temporally consistent, revealing fine hair strands and facial detail with no ghosting.
    }
    \label{fig:motivation}
\end{figure}

Recent advances in dynamic Gaussian splatting~\cite{wu20244d,Jin2024Dyn4d,yang2024deformable,yang2023gs4d,huang2024sc,li2023spacetime,wang2025freetimegs} and monocular dynamic reconstruction~\cite{Li2024MonocularDynamic} model spatio-temporal scenes. In principle, they enable freeze-time synthesis by fixing the temporal coordinate. In practice, frozen renders often show ghosting and blur (Figure~\ref{fig:motivation}). This is common under predominantly forward camera motion as the camera moves away from objects. Existing objectives and benchmarks aim to preserve motion, keeping subjects centered and well supervised in action-focused clips. This favors motion fidelity. In contrast, imperfect Mannequin-Challenge videos are dynamic scene exploration and we focus on artifacts to recover the underlying frozen scene, so a new test set is needed. 

These visual artifacts stem from inconsistent temporal observations. In monocular sequences, Gaussians lack reliable data when they are occluded or move outside the camera's view. When rendering a single frozen instant, inference for these unconstrained Gaussians creates ghosting and blur, corrupting the final reconstruction~\cite{Levoy1996LightField}.

We propose Splannequin, a dual-detection regularization method that stabilizes dynamic Gaussians for freeze-time rendering. It identifies problematic Gaussians as \textbf{hidden} or \textbf{defective}, anchoring hidden ones to their well-observed past states and defective ones to future states with stronger supervision. Implemented as simple loss terms without architectural changes, Splannequin integrates into existing dynamic Gaussian pipelines with minimal overhead. This enables cleaner static views from single-camera captures and improves tolerance to small motions for consumer media and VR/AR tours.

We validate Splannequin on a new benchmark of 10 real-world Mannequin Challenge videos, demonstrating strong gains over state-of-the-art baselines. Our method improves compositional quality assessment (CQA) by up to 243.8\% and technical quality (COVER) by 339.85\% when applied to D-3DGS~\cite{yang2024deformable}. Our contributions are:
\begin{itemize}[leftmargin=*]
    \item \textbf{A Novel Problem Formulation and Benchmark.} We are the first to formally address synthesizing high-fidelity, freeze-time videos from monocular MC footage, providing a new benchmark and evaluation protocol.
    \item \textbf{A Targeted Regularization Framework.} We propose a novel method to identify and regularize \textit{hidden} and \textit{defective} Gaussians, the primary sources of temporal artifacts, and anchor them to reliable past or future states.
    \item \textbf{State-of-the-art performance with zero inference overhead.} We improve visual quality and stability in existing methods without architectural changes. As the deformation runs only once for a target instant, we achieve inference speeds exceeding 280 FPS on an RTX 4090.
\end{itemize}

\vspace{-1mm}
\section{Related Work}
\label{sec:related}
\vspace{-1mm}

The \textsc{Mannequin-Challenge} dataset popularized "frozen-people" clips for depth learning~\cite{li2019learning}, establishing the foundational benchmark for synthesizing freeze-time effects from casual captures. Modern dynamic human renderers such as HumanNeRF~\cite{weng2022humannerf}, HumanNeRF-SE~\cite{ma2024humannerf}, FlexNeRF~\cite{jayasundara2023flexnerf}, FloRen~\cite{shao2022floren}, and Holoported-Characters~\cite{shetty2024holoported} reproduce articulated motion rather than suppress it. Scene motion-aware IBR methods like DynIBaR~\cite{li2023dynibar} and the recent BTimer feed-forward bullet-time pipeline~\cite{liang2024feed,sudmann2016bullet} generalize to arbitrary camera paths but retain micro-motions, with BTimer being the first to use 3D Gaussian Splatting for motion-aware bullet-time synthesis. Shape of Motion~\cite{wang2024shape} addresses fast motion and occlusions in monocular 4D reconstruction relevant to imperfect Mannequin Challenge scenarios. Diffusion-based pipelines (4Real~\cite{yu20244real}, CAT4D~\cite{wu2024cat4d}, Deblur-Avatar~\cite{luo2025deblur}, GAUSSIANFlow~\cite{gao2024gaussianflow}, \cite{Chan2023Gnv}) hallucinate photorealistic 4D geometry yet model sub-frame dynamics. We detect and regularize poorly supervised Gaussians\cite{hou20253d} to erase erroneous placements and appearances while preserving rendering quality.

\mypara{NeRFs and Gaussian Splatting for Static Novel-View Synthesis.}
Neural Radiance Fields (NeRF)\cite{Biswas2024Tfsnerf} model radiance using MLPs\cite{mildenhall2021nerf,barron2021mip,barron2022mip}, while hash-grid encodings reduce training time~\cite{muller2022instant}. 3D Gaussian Splatting achieves real-time rendering~\cite{kerbl20233d}, with recent developments surveyed in~\cite{wu2024recent}. Extensions remove SfM preprocessing (COLMAP-Free 3DGS~\cite{fu2024colmap}), refine camera poses (Look Gauss, No Pose~\cite{schmidt2024look}, LongSplat~\cite{lin2025longsplat}, and joint pose-radiance optimization~\cite{chen2024improving}), and incorporate directional information (6DGS~\cite{gao20246dgs}). To handle large-scale casual captures, progressive optimization significantly improves robustness~\cite{meuleman2023progressively}. Recent artifact removal methods include EFA-GS~\cite{wang2025efa} which eliminates floating artifacts through frequency-domain analysis, VRSplat~\cite{tu2025vrsplat} which addresses temporal popping during viewpoint changes, and 3DGS-HD~\cite{sun20243dgs} which removes unrealistic artifacts. Sparse-view robustness is enhanced by depth priors~\cite{kumar2024few,li2024dngaussian}, multi-view priors (BoostMVSNeRFs~\cite{su2024boostmvsnerfs}, \cite{xu2024mvpgs, Wang2024Mvs}), co-regularization\cite{zhang2024corgs}, neural field integration~\cite{mihajlovic2024splatfields}, structural dropout~\cite{chen2025dropgaussian}, and few-shot convergence strategies without learned priors (FrugalNeRF~\cite{lin2025frugalnerf}). Techniques like BOGausS prune redundant splats~\cite{pateux2025bogauss}, while 2DGS uses oriented disks for improved fidelity~\cite{huang20242d}. Because these static pipelines assume \emph{fully rigid} scenes, minor dynamics cause issues, making new solutions necessary.

\mypara{Dynamic NeRF and Gaussian Representations.}
Early dynamic NeRFs embed time as an additional coordinate or latent code: D-NeRF models a deformation field~\cite{pumarola2021d}, Nerfies incorporates a deformation MLP~\cite{park2021nerfies}, HyperNeRF lifts the canonical space~\cite{park2021hypernerf}, and NR-NeRF targets non-rigid objects~\cite{tretschk2021non}. HexPlane provides foundational six-plane decomposition for 4D spacetime~\cite{cao2023hexplane}. To scale to longer sequences, TiNeuVox employs a time-aware voxel grid~\cite{fang2022fast}, NeRFPlayer streams sub-fields~\cite{song2023nerfplayer}, and RoDynRF refines poses with static and dynamic fields~\cite{liu2023robust}. On the explicit side, 4D Gaussian Splatting enables real-time rendering of deformations~\cite{wu20244d, fan2025spectromotion}, with recent advances including uncertainty-aware regularization for handling poorly supervised regions~\cite{kim2024uncertainty}, spatial-temporal consistency through motion-aware regularization~\cite{li2024st4dgs}, deblurring from blurry monocular videos~\cite{wu2024deblur4dgs}, and depth-prior integration for casual captures (MoDGS~\cite{liu2024modgs}). MotionGS explores explicit motion guidance~\cite{wu2024motiongs}, while Grid4D uses 4D decomposed hash encoding~\cite{jang2024grid4d}. Efficiency improvements include GaussianVideo~\cite{bond2025gaussianvideo}, GIFStream~\cite{li2025gifstream}, VeGaS~\cite{smolak2024vegas}, DynMF's compact motion basis~\cite{kratimenos2024dynmf} and Animatable 3D Gaussians at 60 fps~\cite{ye2023animatable}. Recent benchmarking~\cite{liang2024monocular} reveals the brittleness of monocular dynamic reconstruction. Crucially, all aim to preserve observed motion. In contrast, our \emph{near-static} setting addresses ghosting~\cite{Wang2024TemporallyConsistentGS} by optimizing Gaussians for improved rigid scene and novel-view reconstruction.

\mypara{Video Stabilization and Temporal Coherence.}
2D stabilizers smooth footage via mesh warps. Bundled Camera Paths~\cite{Roberto2022Vstab} optimize homographies, SteadyFlow enforces smooth optical flow~\cite{liu2013bundled,liu2014steadyflow}, and gyroscope filtering corrects rolling-shutter distortion~\cite{bell2014non}. Learning-based methods re-render frames or predict warp fields~\cite{liu2021hybrid,yu2020learning}. Causal variants reduce latency~\cite{zhang2023minimum,shi2022deep}, while depth-aware schemes reconstruct geometry~\cite{lee20213d}. Recent work addresses spatiotemporally inconsistent observations through residual compensation~\cite{zhang2025compensating} and exposure completion for temporal consistency~\cite{cui2024exposure}. LeanVAE provides efficient reconstruction for temporal coherence~\cite{cheng2025leanvae}. A recent survey highlights challenges like border completion~\cite{roberto2022survey}. These output warped 2D frames. Our pipeline unifies \emph{3D} stabilization and novel-view synthesis~\cite{sajjadi2022scene,Berian2025Nvs} via Gaussian splatting, removing Gaussian artifacts~\cite{Zhang2023GaussianTracking} while enabling free-viewpoint playback~\cite{Zitnick2004High-QualityVS}.

\mypara{Micro-Motion Detection and Repair.}
Eulerian video magnification reveals imperceptible shifts by amplifying per-pixel traces~\cite{wu2012eulerian}, with phase-based variants improving robustness~\cite{wadhwa2013phase}. Methods handle large displacements through layer-wise alignment~\cite{elgharib2015video}, learned filters~\cite{oh2018learning}, and Transformer-based denoising~\cite{wang2024eulermormer}. All remain 2D and \emph{amplify} motion. In 3D, Feng \etal\ model subtle dynamics via time-varying radiance fields~\cite{feng20233d}, while early geometry-aware editing hinted at sub-pixel motion control~\cite{kumar20003d}. Recent work handles unobserved views through refinement and fusion~\cite{liu2022refu}, and applies static restoration priors for dynamic regularization~\cite{chen2025rsrnf}. Recent reconstructions capture fine dynamics without suppression~\cite{wang2024shape}. We operate \emph{within} the Gaussian representation: a confidence-weighted classifier identifies and improves \emph{hidden} and \emph{defective} primitives, removing sub-pixel jitter and enabling robust freeze-time synthesis in real-world, near-static scenes.
\vspace{-1mm}
\section{Background}
\label{sec:background}
\vspace{-1mm}

Dynamic Gaussian splatting extends 3DGS to model time-varying scenes. Given a set of $N$ images, camera rotations and translations, and timestamps, $\{(I_n, \mathbf{R}_n, \mathbf{b}_n, t_n)\}_{n=1}^N$, the scene is represented by a set of canonical 3D Gaussians $\{G_k\}_{k=1}^K$. Each Gaussian $G_k$ is defined by a static mean $\boldsymbol{\mu}_k$ and covariance $\boldsymbol{\Sigma}_k$. A deformation network $f_\theta$, typically an MLP, predicts the temporal evolution of each Gaussian:
\begin{equation}
(\Delta \boldsymbol{\mu}_{k,t}, \Delta \boldsymbol{\Sigma}_{k,t}) = f_\theta(\boldsymbol{\mu}_k, t).
\end{equation}
The time-dependent state of the Gaussian is then $G_k(t) = (\boldsymbol{\mu}_k + \Delta \boldsymbol{\mu}_{k,t}, \boldsymbol{\Sigma}_k + \Delta \boldsymbol{\Sigma}_{k,t})$. The model is optimized by minimizing a photometric loss between the rendered image $\hat{I}_n$ and the ground truth image $I_n$ for each observation:
\begin{equation}
\mathcal{L}_{\text{recon}} = \sum_{n=1}^N \ell(\hat{I}(\mathbf{R}_n, \mathbf{b}_n, t_n), I_n).
\end{equation}
This framework enables high-fidelity modeling of dynamic scenes. However, as we will detail, its reliance on direct supervision at each timestamp makes it prone to artifacts when applied to our specific problem.

\vspace{-1mm}
\section{Problem Definition}
\label{sec:our_prob}
\vspace{-1mm}

We address the problem of frozen scene reconstruction from a monocular Mannequin Challenge (MC) style video. The goal is to synthesize a high-fidelity static rendering $\hat{I}(\mR, \vb, t^\star)$ from any input camera view $(\mR, \vb)$ at a single, user-selected timestamp $t^\star$, effectively freezing all motion.

The dynamic framework is a natural fit, as rendering a view of an instant seems as simple as fixing the time parameter to $t^\star$. However, this fails for monocular MC-style videos. The reconstruction loss, $\mathcal{L}_{\text{recon}}$, only provides a training signal for a Gaussian $G_k$ at timestamps $t_n$ where it is visible and receives gradients in the image $I_n$. The Gaussian is also unobserved at time $t_n$ (due to occlusion or leaving the camera frustum), and its deformation $f_\theta(\boldsymbol{\mu}_k, t_n)$ receives no gradient. This sparse supervision leads to unstable state estimates at unobserved times.

\begin{figure}
    \centering
    \vspace{-3mm}
    \includegraphics[width=0.97\linewidth]{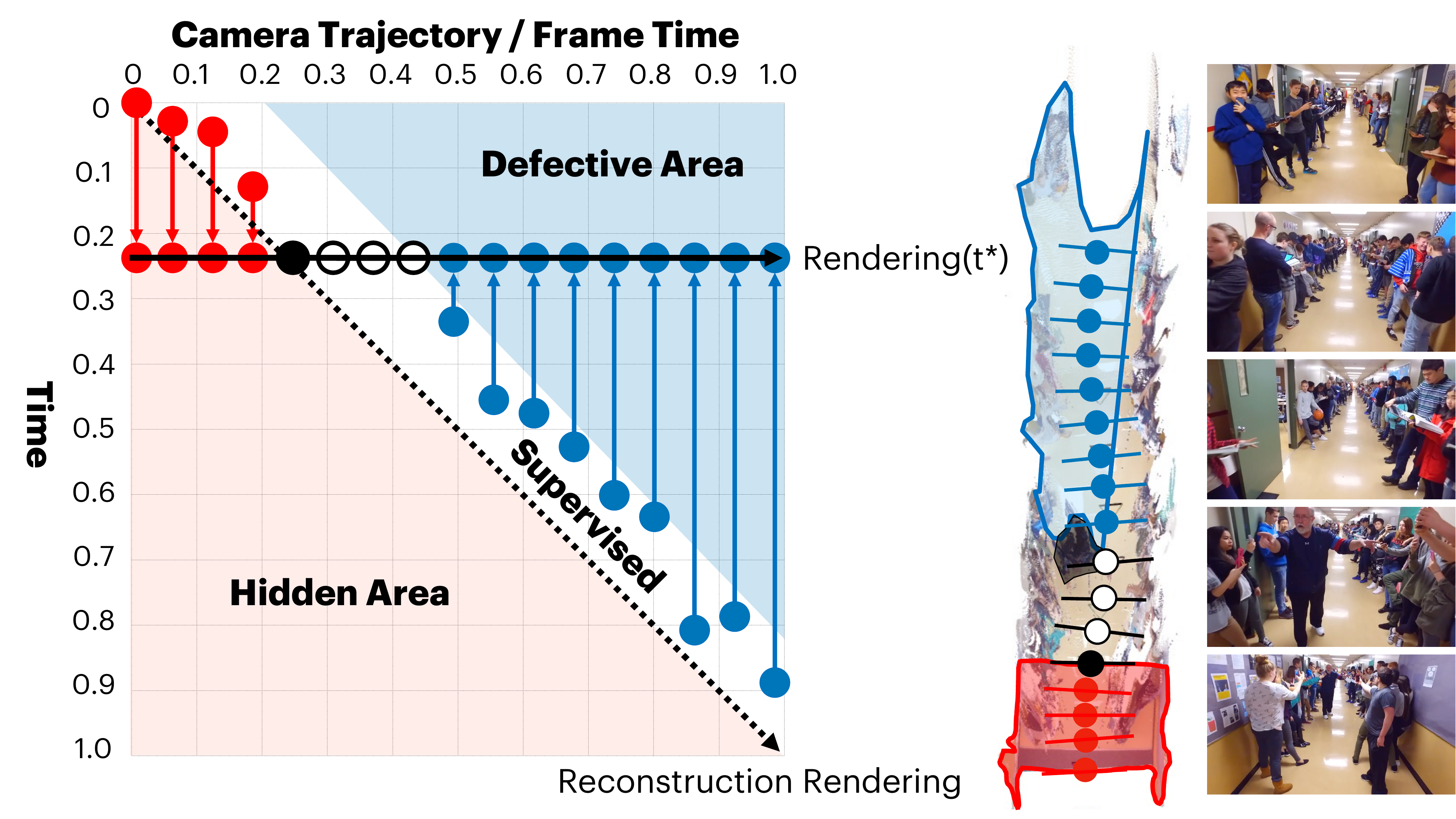}
    \vspace{-3mm}
    \caption{\textbf{Time-Camera Conceptualization.} Assuming forward camera motion, the diagonal dashed line represents standard dynamic rendering, while the horizontal line shows freeze-time rendering at a fixed timestamp \(t^\star\). Along this freeze-time line, unsupervised Gaussians are either \textbf{hidden} (red points, as the camera has passed them) or \textbf{defective} (blue points, not yet well-observed). Our approach regularizes these problematic Gaussians by anchoring them to their supervised counterparts from other timestamps: hidden (red) Gaussians use past states, and defective (blue) Gaussians use future states. The right panel shows a bird's-eye view of a hallway, illustrating how the camera's path creates defective and hidden regions.}
    \label{fig:time_plane}
\end{figure}

To analyze this supervision gap, we use a space-time diagram (Figure~\ref{fig:time_plane}) that plots time against the camera's spatial path. The video, or the standard dynamic reconstruction, is then captured along a diagonal trajectory, but the desired freeze-time render lies along a horizontal line at $t^\star$. This visualization reveals that artifacts arise directly from rendering primitives at timestamps when they were ill-supervised.

\vspace{-1mm}
\subsection{Identification of Ill-supervised Gaussians}
\label{sec:identification}
\vspace{-1mm}

\begin{figure}[t]
    \centering
    \vspace{-3mm}
    \includegraphics[width=0.97\linewidth]{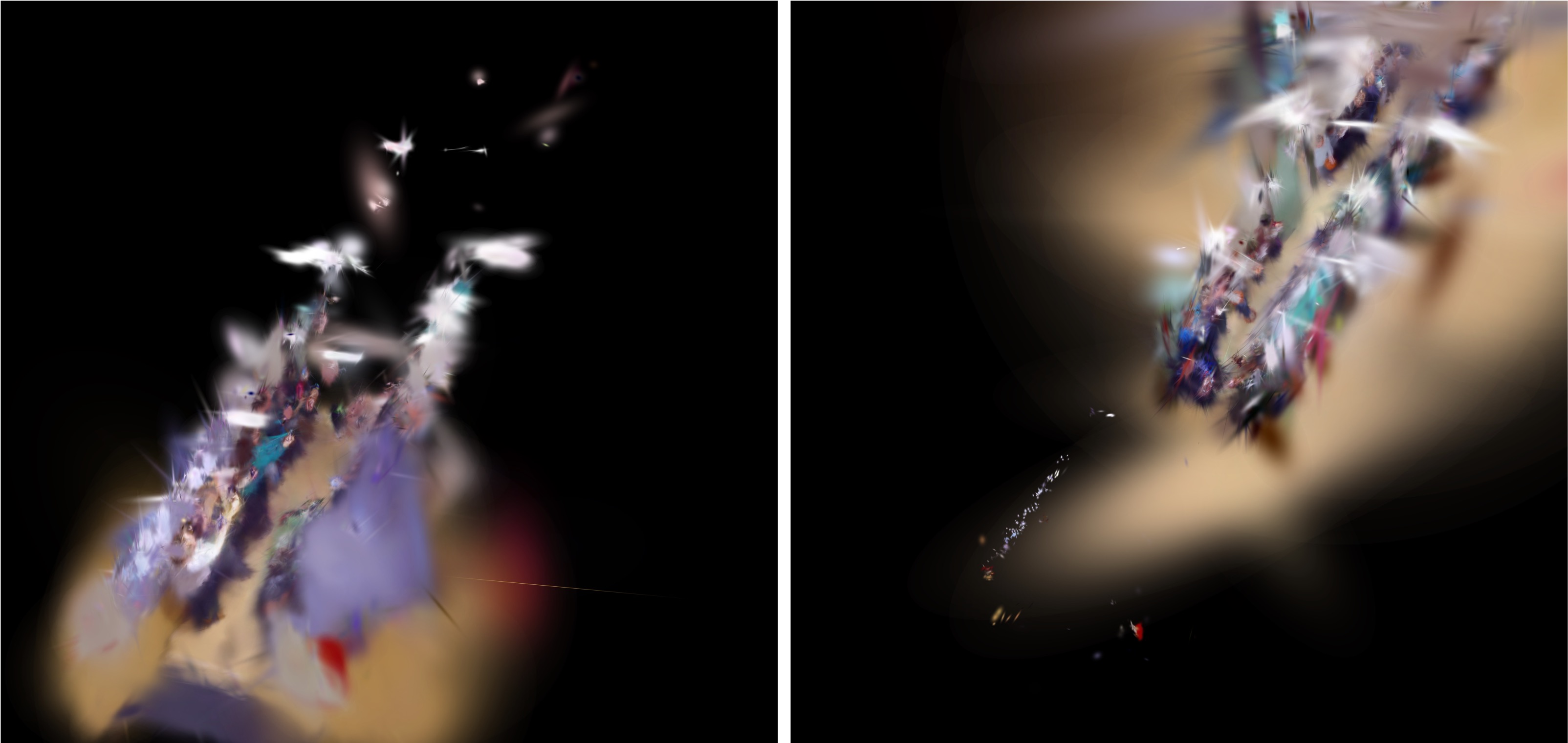}
    \caption{\textbf{Illustration of hidden Gaussians.} Given a timestamp, hidden Gaussians (Left) lie outside the camera frustum, receiving no supervision, while \textbf{visible} Gaussians (Right) are rasterized to form the image. Our method targets ill-supervised hidden Gaussians to prevent visual artifacts.}
    \label{fig:visualization}
\end{figure}

In dynamic Gaussian splatting, primitives only receive supervision from the reconstruction loss when they contribute to the rendered image. We formally identify two failure cases of missing supervision for a Gaussian \(k\) at time $t$. Our regularization strategy identifies these cases and applies targeted temporal anchoring (Figure~\ref{fig:visualization}).

\begin{itemize}[leftmargin=*]
    \item \textbf{Hidden Gaussian:} A primitive \(k\) at time \(t\) is \emph{hidden} if the primitive center projects outside the camera frustum, typically after the camera has moved past it. Such primitives are unobservable and receive no supervision, defined as $s_{\text{hidden}}(k, t)$ = \{1, visibility is 0; 0, otherwise\}.
    
    \item \textbf{Defective Gaussian:} A primitive \(k\) at time \(t\) is \emph{defective} if its center is within the camera frustum while the rendered contribution is negligible, resulting in a zero-gradient update, defined as $s_{\text{defective}}(k, t)$ = \{1, visibility is 1 with gradient $\leq$ 1e-9; 0, otherwise\}. This often happens for distant, highly transparent, or occluded primitives.
\end{itemize}

A primitive \(k\) at time $t$ is therefore \emph{well-supervised} only if $s_{\text{hidden}}(k, t) = 0$ and $s_{\text{defective}}(k, t) = 0$. In practice, we get visibility from the differentiable rasterizer outputs~\cite{kerbl3Dgaussians}.

\vspace{-1mm}
\section{Approach}
\label{sec:approach}
\vspace{-1mm}

\begin{figure*}
    \centering
    \vspace{-3mm}
    \includegraphics[width=0.97\textwidth]{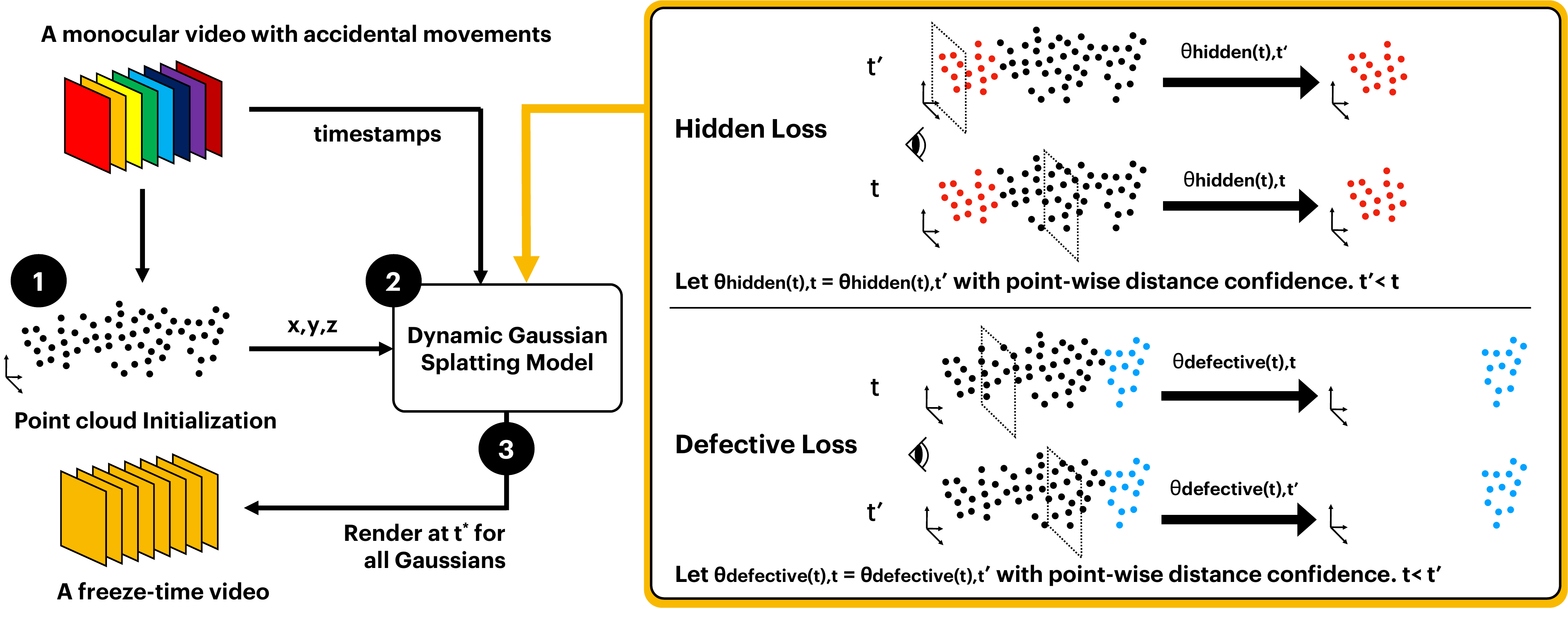}
  \vspace{-3mm}
    \caption{\textbf{Splannequin Pipeline Overview}. The pipeline: (1) extracts point clouds from input video, (2) use dynamic Gaussian splatting with dual-detection losses that anchor hidden Gaussians to earlier frames (t' $<$ t) and defective Gaussians to later frames (t $<$ t'), and (3) renders freeze-time videos at any timestamp $t^\star$. Temporal distance-based confidence weighting ensures appropriate regularization strength, with closer reference frames providing stronger anchoring than distant ones for robust temporal consistency and artifact elimination.}
    \label{fig:pipeline}
\end{figure*}

To recover a frozen scene from an MC-style video, we address the parameter drift of ill-supervised Gaussians. Splannequin is a regularization framework that identifies and stabilizes these Gaussians (Figure~\ref{fig:pipeline}). We classify each ill-supervised Gaussian at a given timestamp as either hidden or defective. We then apply a targeted temporal consistency loss that anchors the parameters of each ill-supervised Gaussian to a nearby, well-supervised reference state, weighted by temporal distance to respect genuine motion while suppressing artifacts.

\vspace{-1mm}
\subsection{Temporally-Anchored Regularization}
\label{sec:regularization}
\vspace{-1mm}
For each ill-supervised Gaussian, we enforce temporal consistency by regularizing its parameters $\vtheta_k(t)$ toward a stable reference state, including position $\mu_k(t)$, covariance $\boldsymbol{\Sigma}_k(t)$, opacity $\alpha_k(t)$, and spherical harmonic coefficients c. During each training iteration, for an ill-supervised Gaussian $k$ at time $t$, we randomly select a timestamp $t_{\text{ref}}$, described in the next paragraphs, from the set of all timestamps and check if it is a \emph{well-supervised} state. This avoids creating explicit anchor pools or performing expensive searches.

\subsubsection{Anchoring Hidden Gaussians}
If a Gaussian $k$ at time $t$ is \emph{hidden} ($s_{\text{hidden}}(k, t) = 1$), its state should be ideally constrained by its recent appearance. In practice, a randomly sampled reference time $t_{\text{ref}}$ is a valid anchor if it is in the past ($t_{\text{ref}} < t$) and the Gaussian $k$ is well-supervised at $t_{\text{ref}}$. When these conditions are met, we apply the consistency loss defined as:
\begin{equation}
\mathcal{L}_{\text{consistency}}(k, t) = \phi(t, t_{\text{ref}}) \cdot \mathcal{D}(\vtheta_k(t), \vtheta_k(t_{\text{ref}})),
\end{equation}
where the discrepancy measure $\mathcal{D}$ is the $L_1$ or $L_2$ norm:
\begin{equation}
\mathcal{D}(\vtheta_{A}, \vtheta_{B}) =
\begin{cases}
    \|\vtheta_{A} - \vtheta_{B}\|_1, & \text{for L1 norm} \\
    \|\vtheta_{A} - \vtheta_{B}\|_2^2, & \text{for L2 norm}
\end{cases}
\end{equation}
The term $\phi(t, t_{\text{ref}})$ is weighting. For well-supervised anchors, we conservatively down-weight their influence using an exponential decay based on temporal distance, $\phi = e^{-\tau |t - t_{\text{ref}}|}$.

\subsubsection{Anchoring Defective Gaussians}
Conversely, if a Gaussian $k$ at time $t$ is \emph{defective} ($s_{\text{defective}}(k, t) = 1$), it typically lacks supervision because the camera has not yet reached it. In practice, a randomly sampled reference time $t_{\text{ref}}$ is a valid anchor only if it is in the future ($t_{\text{ref}} > t$) and the Gaussian $k$ is well-supervised at $t_{\text{ref}}$. If so, we apply the same consistency loss.

\subsubsection{Total Objective Function}
The final training objective combines the standard reconstruction loss with our regularization terms, summed over all instances where a valid anchor was found:
\begin{align}
\mathcal{L} = \mathcal{L}_{\text{recon}} 
    & + \lambda_{\text{hidden}}\sum \mathcal{L}_{\text{hidden}}(k, t) \\ 
    & + \lambda_{\text{defective}}\sum \mathcal{L}_{\text{defective}}(k, t).
\end{align}
where $\mathcal{L}_{\text{hidden}}$ and $\mathcal{L}_{\text{defective}}$ are instances of $\mathcal{L}_{\text{consistency}}$, and $\lambda_{\text{hidden}}$ and $\lambda_{\text{defective}}$ are weighting hyperparameters.

\vspace{-1mm}
\subsection{Fixed-Time Rendering}
\vspace{-1mm}
After training, our primary application is fixed-time rendering. A freeze-time video can be synthesized by rendering all $N$ training poses $\{(\mR_n, \vb_n)\}_{n=1}^N$ at a single, user-selected timestamp $t^\star$. Because our method preserves temporal variation, it grants users the flexibility to scrub through time, select the exact desired moment, and generate a perfectly static video.

\vspace{-1mm}
\section{Experiments}
\label{sec:exp}
\vspace{-1mm}

\begin{figure*}
    \centering
    \vspace{-3mm}
    \includegraphics[width=0.965\textwidth]{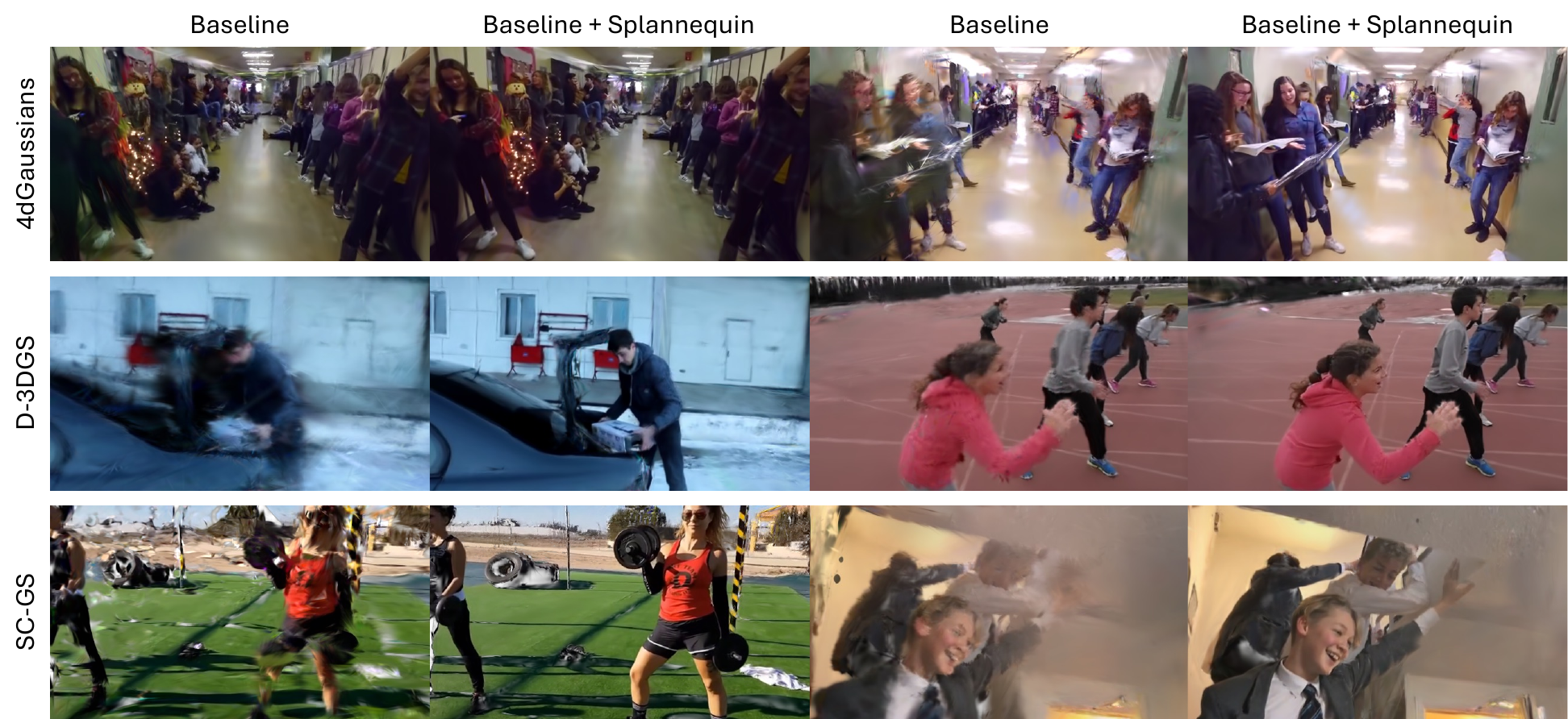}
    \caption{\textbf{Qualitative Comparison across Our Real-World Benchmark.} Each column shows freeze-time renderings from all methods at a viewpoint. Rows correspond to direct comparisons of identical viewpoints with baselines: 4DGaussians (top), D-3DGS (middle), and SC-GS (bottom). Adding Splannequin consistently produces sharper, more temporally coherent results, exhibiting reduced ghosting and artifact suppression compared to baseline methods.}
    \label{fig:eval_real_world}
\end{figure*}
\begin{figure*}
    \centering
    \includegraphics[width=0.97\textwidth]{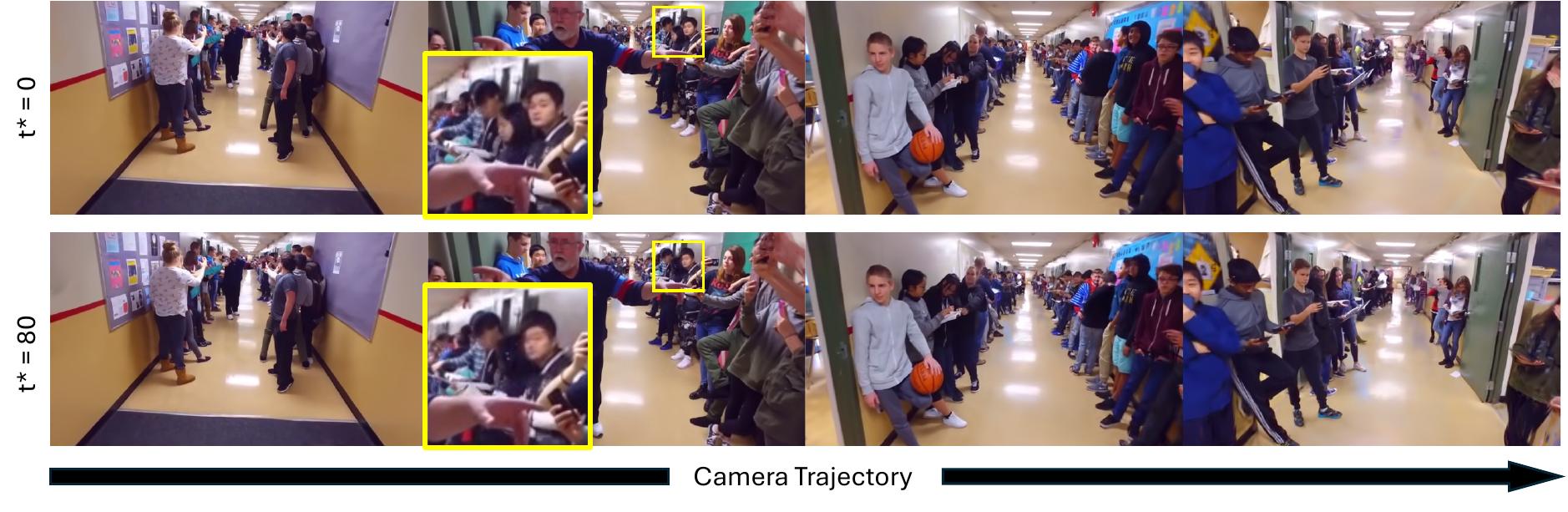}
    \vspace{-3mm}
    \caption{\textbf{User-Selectable Freeze-Time Instants.} Splannequin empowers users to select the precise moment to freeze, allowing for artistic control over the final scene. Both rows show high-fidelity freeze-time videos generated from the same input sequence but frozen at two different, user-selected timestamps. \emph{Top:} At Timestamp 0, the subject in the inset is looking down. \emph{Bottom:} At Timestamp 80, captured seconds later, the subject has turned their head. Our method successfully reconstructs both moments with sharp detail and stability, preserving these subtle differences and enabling creative selection based on pose and expression.}
    \label{fig:user-selection}
\end{figure*}
\begin{figure}
    \centering
    \vspace{-3mm}
    \includegraphics[width=\columnwidth]{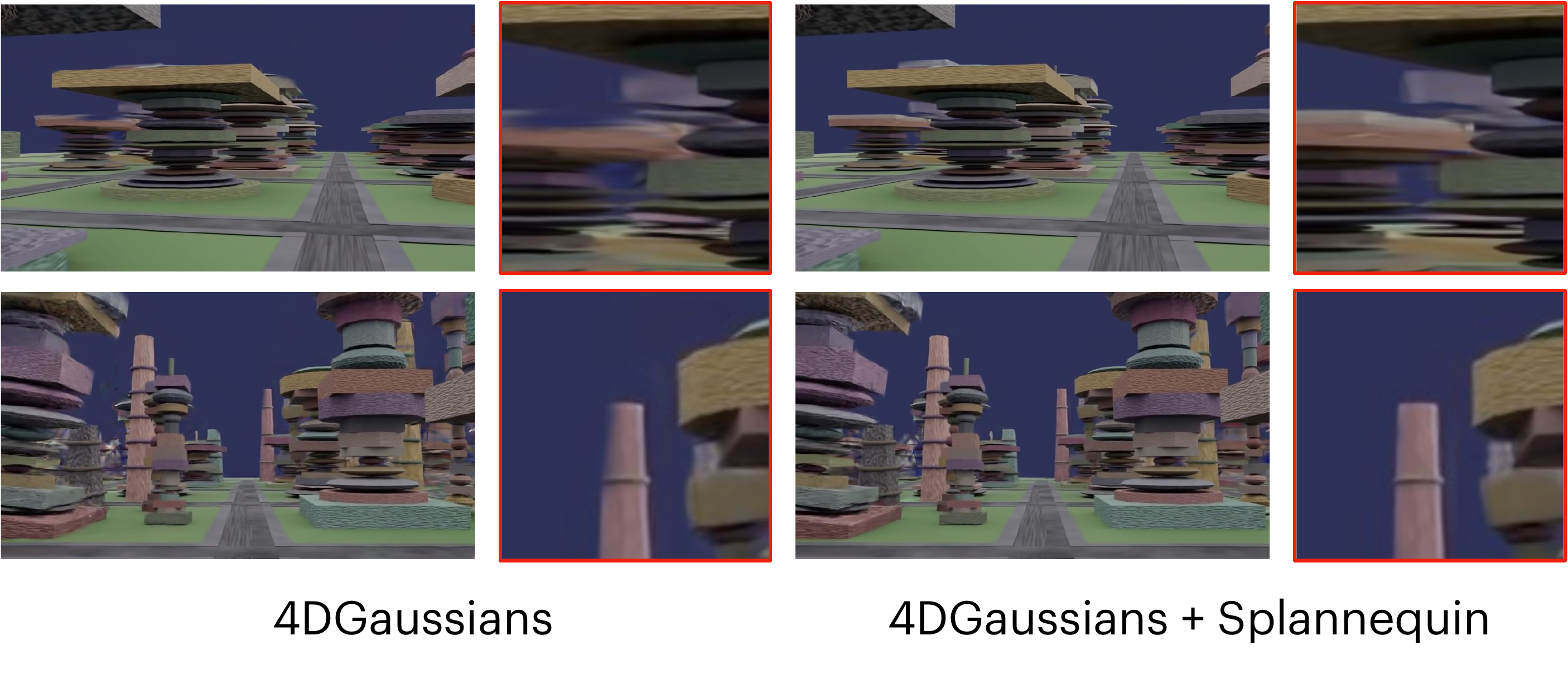}
    \vspace{-7mm}
    \caption{\textbf{Validation on Simulated Dataset.} Qualitative comparison between 4DGaussians (left) and 4DGaussians + Splannequin (right) on synthetic scenes with ground truth. With Splannequin, the results better preserve geometric details and suppressed artifacts, as validated against static reference frames. Insets highlight regions of improved structural fidelity.}
    \label{fig:exp_blender}
\end{figure}


\subsection{Implementation Details}
\vspace{-1mm}
We implemented in PyTorch, extending the official dynamic Gaussian splatting framework, with all experiments run on a single NVIDIA RTX 4090 GPU. We follow standard adaptive densification practices and train each model for 30,000 iterations. 
The regularization is introduced in stages to allow the base geometry to stabilize. Our main losses, \(\mathcal{L}_{\text{hidden}}\) and \(\mathcal{L}_{\text{defective}}\), begin at iteration 10,000, initially using an L2 norm for the discrepancy measure \(\mathcal{D}\) and switching to an L1 norm at iteration 20,000. We use loss weights \(\lambda_{\text{hidden}} = \lambda_{\text{defective}} = 10\) and a confidence decay factor of \(\tau=5\). These regularization losses are computed every 10 iterations by sampling two random view-timestamp pairs and applying the anchoring logic only when valid supervised anchor states are available for the target primitives.

\vspace{-1mm}
\subsection{Experimental Setup}
\vspace{-1mm}
To ensure comprehension and robustness, we test across multiple timestamps for each video. After training a model on a full sequence, we render multiple freeze-time videos by using timestamps from every 8th frame of the input video as the target freeze-time $t^\star$. This process yields a diverse set of ``frozen'' clips per scene for a holistic assessment of temporal consistency and artifact reduction across the entire duration of the original capture.

\vspace{2pt}
\noindent {\bf Benchmark.}
Our evaluation uses a challenging custom dataset of 10 640$\times$360 Mannequin Challenge-style videos (2,869 frames, 361 fixed-time renderings) from the public Google Mannequin Challenge collection~\cite{li2019learning}. Spanning diverse scenes, including seven indoor scenes and three outdoor fields, with subjects exhibiting natural, unscripted micro-motions, the dataset is characterized by sparse temporal supervision. On average, the benchmark has less than 10\% consistent visibility across the full videos.

\vspace{2pt}
\noindent {\bf Metrics.}
Since the ground-truth frames contain the very motions we aim to suppress, we primarily rely on a suite of no-reference (NR) perceptual quality metrics that correlate well with human judgments, rather than full-reference metrics. In all our tables, improvements are reported as relative gains computed as (x - baseline) / baseline $\times$ 100.
\begin{itemize}[leftmargin=*]
    \item \textbf{Composition (CQA)}: Adapted from View Evaluation Net (VEN)~\cite{Wei2018VEN} to assess compositional clarity by comparing rendering quality at identical viewpoints.
    \item \textbf{TOPIQ-NR}~\cite{chen2024topiq}: A unified model leveraging multiple feature types to provide a general-purpose quality score.
    \item \textbf{CLIP-IQA}~\cite{wang2023exploring}: A metric assessing quality by measuring the similarity between rendered image features and textual quality descriptors in CLIP.
    \item \textbf{MUSIQ}~\cite{ke2021musiq}: A multi-scale, Transformer-based model evaluating both fine-grained detail and global quality.
    \item \textbf{HyperIQA}~\cite{Su_2020_CVPR}: An adaptive model generating content-specific weights to assess quality across diverse scenes.
    \item \textbf{COVER}~\cite{he2024cover}: A holistic video quality assessor that evaluates semantic, technical, and aesthetic dimensions.
\end{itemize}

\vspace{2pt}
\noindent {\bf Compared Methods.}
\label{sec:compared_methods}
We integrated Splannequin with three state-of-the-art dynamic Gaussian Splatting methods: \textbf{D-3DGS}~\cite{yang2024deformable}, \textbf{SC-GS}~\cite{huang2024sc}, and \textbf{4DGaussians}~\cite{wu20244d}. To ensure a fair comparison, all baselines were trained under identical conditions using their default hyperparameters. We excluded other recent methods from our primary comparison for specific reasons: MoDGS~\cite{liu2024modgs} relies on pre-computed depth maps that conflate camera and object motion, and STG~\cite{li2023spacetime} requires a multi-camera setup.

\begin{table}[t]
\centering
    \vspace{-3mm}
\caption{\textbf{Validation on our synthetic dataset.} We report absolute scores for reference metrics and percentage improvements for non-referenced quality assessment metrics.}
\vspace{-3mm}
\label{table:synthetic}
\resizebox{\columnwidth}{!}{%
\begin{tabular}{lcccc}
\toprule
\multicolumn{5}{l}{\textit{Referenced Metrics}} \\
\midrule
\textbf{Method} & \textbf{PSNR}\,$\uparrow$ & \textbf{SSIM}\,$\uparrow$ & \textbf{LPIPS}\,$\downarrow$ & \textbf{FVD}\,$\downarrow$ \\
\cmidrule(r){1-1} \cmidrule(l){2-5}
4DGaussians & 28.03 & 0.81 & 0.09 & 98.93 \\
4DGaussians + Splannequin & \textbf{28.85} & \textbf{0.83} & \textbf{0.08} & \textbf{82.73} \\
\midrule
\end{tabular}%
}
\resizebox{\columnwidth}{!}{%
\begin{tabular}{ccccc}
\multicolumn{5}{l}{\textit{Average Image Quality Assessment (IQA) Metric Improvements}} \\
\midrule
\textbf{CQA $\uparrow$} & \textbf{TOPIQ-NR $\uparrow$} & \textbf{CLIP-IQA $\uparrow$} & \textbf{MUSIQ $\uparrow$} & \textbf{HyperIQA $\uparrow$} \\
\cmidrule(l){1-5}
26.43\% & 2.08\% & 1.18\% & 1.74\% & 2.13\% \\
\midrule
\multicolumn{5}{l}{\textit{Video Quality Assessment (VQA) Metric Improvements}} \\
\midrule
\multirow{2}{*}{\textbf{COVER $\uparrow$}} & \textbf{Semantic} & \textbf{Technical} & \textbf{Aesthetic} & \textbf{Overall} \\
\cmidrule(l){2-5}
& 1.29\% & 3.11\% & 95.60\% & 6.56\% \\
\bottomrule
\end{tabular}%
}
\end{table}

\begin{table}[t]
\centering
\caption{\textbf{Quantitative comparison on our real-world dataset.} The values represent the percentage improvement Splannequin provides when added to each baseline method (higher is better). Our method consistently enhances all baselines, with the most gains in technical artifact suppression (COVER Technical) and on the lowest-quality frames (IQA Bottom 25\%). Methods are abbreviated as: (1) 4DGaussians+, (2) D-3DGS+, and (3) SC-GS+. W.F. is the worst frame.}
\vspace{-3mm}
\label{table:real_world_final}
\resizebox{\columnwidth}{!}{%
\begin{tabular}{llccccc}
\toprule
\multicolumn{7}{l}{\textit{Video Quality Assessment (VQA) Metric Improvement}} \\
\midrule
\cmidrule(lr){4-6}
\multicolumn{2}{c}{\textbf{Metric / Method}} & \textbf{Semantic} & \textbf{Technical} & \textbf{Aesthetic} & \textbf{Overall} & \\
\midrule
\multirow{3}{*}{\textbf{COVER}}
& (1) & 2.23\% & 73.03\% & 20.53\% & 68.25\% & \\
& (2) & 2.97\% & 339.85\% & 75.24\% & 183.68\% & \\
& (3) & 1.84\% & 81.53\% & 30.46\% & 121.70\% & \\
\midrule
\multicolumn{7}{l}{\textit{Image Quality Assessment (IQA) Metric Improvement}} \\
\cmidrule(lr){1-7}
\multicolumn{2}{c}{\multirow{2}{*}{\textbf{Metric / Method}}} & \multirow{2}{*}{\textbf{Average}} & \multicolumn{3}{c}{\textbf{Bottom Percentage}} & \multirow{2}{*}{\textbf{W.F.}} \\
\cmidrule(lr){4-6}
& & & \textbf{75\%} & \textbf{50\%} & \textbf{25\%} & \\
\midrule
\multirow{3}{*}{\textbf{CQA}}
& (1) & 121.33\% & 60.97\% & 51.54\% & 101.44\% & 18.48\% \\
& (2) & 243.80\% & 65.94\% & 103.11\% & 98.64\% & 17.39\% \\
& (3) & 48.88\% & 38.45\% & 404.08\% & 36.31\% & 26.11\% \\
\midrule
\multirow{3}{*}{\textbf{TOPIQ-NR}}
& (1) & 2.53\% & 3.61\% & 4.98\% & 6.76\% & 13.65\% \\
& (2) & 7.10\% & 9.78\% & 13.40\% & 17.82\% & 28.35\% \\
& (3) & 8.26\% & 11.87\% & 16.20\% & 22.62\% & 47.81\% \\
\midrule
\multirow{3}{*}{\textbf{CLIP-IQA}}
& (1) & 2.42\% & 3.70\% & 5.27\% & 7.44\% & 15.99\% \\
& (2) & 6.96\% & 8.99\% & 11.51\% & 14.46\% & 24.21\% \\
& (3) & 8.72\% & 12.61\% & 16.37\% & 20.19\% & 34.43\% \\
\midrule
\multirow{3}{*}{\textbf{MUSIQ}}
& (1) & 1.29\% & 1.82\% & 2.49\% & 3.51\% & 8.75\% \\
& (2) & 6.62\% & 9.23\% & 13.10\% & 18.60\% & 30.45\% \\
& (3) & 6.43\% & 9.22\% & 12.92\% & 18.50\% & 35.75\% \\
\midrule
\multirow{3}{*}{\textbf{HyperIQA}} 
& (1) & 4.60\% & 5.79\% & 7.23\% & 9.01\% & 12.12\% \\
& (2) & 7.14\% & 9.53\% & 12.04\% & 15.60\% & 21.33\% \\
& (3) & 5.90\% & 8.30\% & 10.88\% & 14.82\% & 30.50\% \\
\bottomrule
\end{tabular}%
}
\end{table}

\begin{table*}[t]
\centering
\caption{\textbf{Ablation study.} We report the percentage degradation in performance when removing the hidden loss and the defective loss individually. The results show that both components are critical.}
\vspace{-3mm}
\label{table:FTVA-ablation}
\resizebox{\textwidth}{!}{%
\begin{tabular}{lccccccccc}
\toprule
\multirow{2}{*}{\textbf{Method}} & \multicolumn{4}{c}{\textbf{COVER} \textit{(VQA Metrics)}} & \multicolumn{5}{c}{\textbf{ Average} \textit{(IQA Metrics)} }\\
\cmidrule(lr){2-5} \cmidrule(lr){6-10}
& \textbf{Semantic} & \textbf{Technical} & \textbf{Aesthetic} & \textbf{Overall} & \textbf{CQA} & \textbf{TOPIQ-NR} & \textbf{CLIP-IQA} & \textbf{MUSIQ} & \textbf{HyperIQA} \\
\midrule
No Hidden Loss & -2.27\% & -94.22\% & -42.45\% & -1072.41\%  & -162.23\% & -3.02\% &  -2.65\% &  -2.33\% &  -4.31\% \\
No Defective Loss & -4.79\% & -197.87\% & -56.25\% & -1027.33\% & -779.13\% & -12.52\% & -4.05\% & -13.32\% & -10.21\% \\
\bottomrule
\end{tabular}%
}
\end{table*}

\vspace{-1mm}
\subsection{Validation on Simulated Benchmark}
\vspace{-1mm}
To validate our approach in a controlled environment, we first tested it on a synthetic dataset of 10 Blender-generated scenes (2,400 frames, 300 fixed-time renderings) with known static ground truth, allowing for reference-based evaluation. As shown in Table~\ref{table:synthetic} and Figure~\ref{fig:exp_blender}, applying Splannequin to the 4DGaussians yielded consistent improvements across all reference metrics. These results verify that Splannequin successfully regularizes artifacts for reconstructions measurably closer to the ideal static geometry while preserving motions, providing a strong foundation for our primary evaluation on real-world data.

\vspace{-1mm}
\subsection{Real-World Evaluation}
\vspace{-1mm}

As an architecture-agnostic framework, Splannequin integrates into existing pipelines with substantial improvements (Figure~\ref{fig:eval_real_world}). Using temporal anchoring (Section~\ref{sec:our_prob}), it achieves large gains over baselines (Table~\ref{table:real_world_final}) while maintaining high efficiency at over 280 FPS, as deformation only needs to be run once for a timestamp. Interestingly, improvement patterns reflect anchor availability: frames with the worst CQA scores suffer from compositional constraints tied to viewpoint rather than temporal supervision, yielding smaller relative gains than the overall average. In contrast, frames with the worst perceptual scores, which directly reflect artifacts, benefit more from temporal anchoring.

Although 4DGaussians + Splannequin (4DGaussians+) achieves performance comparable to static 3DGS reconstruction (Table~\ref{table:3dgs-compare}), our approach uniquely enables motion preservation and user-controlled timestamp selection that 3DGS cannot provide. This allows users to select the precise instant to freeze and preserve subtle artistic details like a subject's pose or expression, as shown in Figure~\ref{fig:user-selection}.

\vspace{-1mm}
\subsection{Ablation Studies}
\label{sec:ablation}
\vspace{-1mm}

\noindent {\bf Effect of Regularization Losses.}
In Table~\ref{table:FTVA-ablation}, removing either component results in a performance drop as they play distinct and complementary roles. Removing the \emph{hidden loss} indicates that anchoring past, out-of-view primitives maintains overall scene stability. In contrast, removing the \emph{defective loss} suggests that anchoring future, not-yet-clear primitives reconstructs the initial geometry and composition of the scene. Together, these results confirm that both regularization terms are critical.

\begin{figure}
    \centering
    \includegraphics[width=\linewidth]{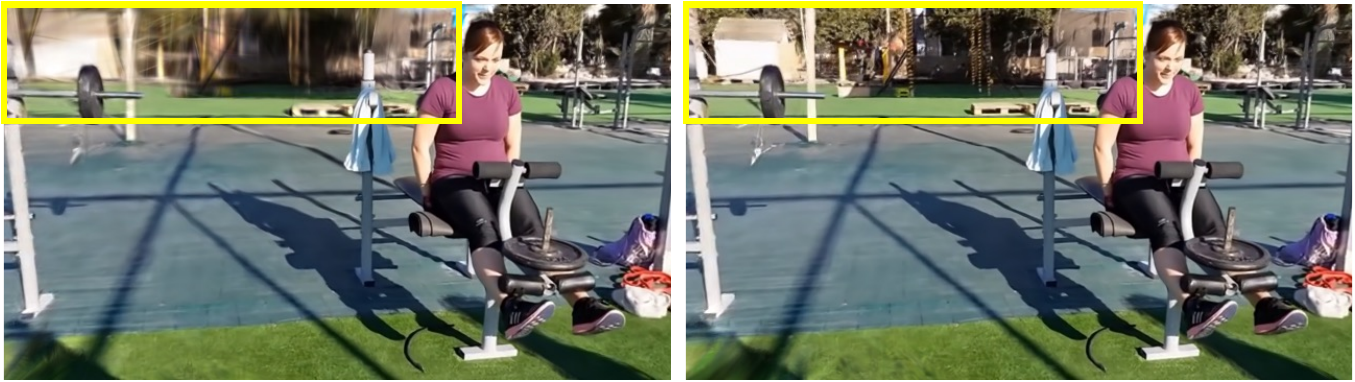}
    \vspace{-6mm}
    \caption{\textbf{Effect of Confidence Weighting.} Visualization comparing results with (right) and without (left) confidence weighting. Without confidence, regularization can over-smooth the frame.}
    \label{fig:ablation-component}
\end{figure}

\vspace{2pt}
\noindent {\bf Effect of Confidence Weighting.}
Figure~\ref{fig:ablation-component} shows that the confidence weighting plays a role in preserving overall fidelity, whereas the unweighted version can over-regularize and blur frames.

\begin{figure}
    \centering
    \includegraphics[width=\linewidth]{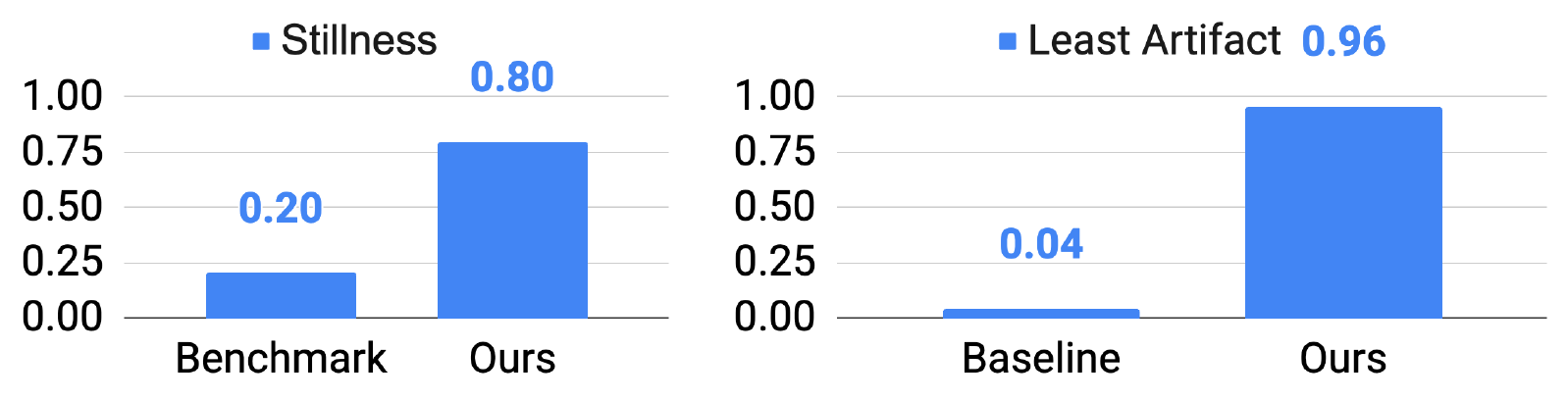}
    \caption{\textbf{User study results.} Our method was preferred in 96\% of comparisons for better visual appeal and fewer artifacts. 80\% of our results were perceived as more "perfectly frozen" than the original captures, validating our approach.}
    \label{fig:userstudy}
    
\end{figure}


\begin{table}[t]
\centering
\caption{\textbf{Comparison to 3DGS.} 4DGaussians+ is comparable to 3DGS when treating scenes as static. However, 3DGS cannot preserve motions or allow per-timestamp video evaluation.}
\vspace{-3mm}
\label{table:3dgs-compare}
\resizebox{\columnwidth}{!}{%
\begin{tabular}{llccccc}
\toprule
\multirow{2}{*}{\textbf{Metric}} & \multirow{2}{*}{\textbf{Average}} & \multicolumn{3}{c}{\textbf{Bottom Percentage}} & \multirow{2}{*}{\textbf{W.F.}} \\
\cmidrule(lr){3-5}
& & \textbf{75\%} & \textbf{50\%} & \textbf{25\%} & \\
\midrule
\textbf{CQA}
& 23.18\% & 46.31\% & 27.80\% & 22.09\% & 4.77\% \\
\textbf{TOPIQ-NR}
& -0.11\% & 0.77\% & 1.89\% & 3.80\% & 20.68\% \\
\textbf{CLIP-IQA}
& -2.17\% & -0.33\% & 1.73\% & 3.85\% & 16.95\% \\
\textbf{MUSIQ}
& -0.59\% & 0.10\% & 0.89\% & 2.64\% & 11.38\% \\
\textbf{HyperIQA}
& 2.95\% & 3.89\% & 5.33\% & 7.46\% & 14.14\% \\
\bottomrule
\end{tabular}%
}
\end{table}

\vspace{2pt}
\noindent {\bf User Study.}
The ultimate goal of this work is to create a convincing ``freeze-time illusion.'' We conducted a user study where 23 participants judged videos on visual quality and perceived \textbf{stillness}. Our method was strongly preferred, with 96\% preference for fewer artifacts and better visual appeal (Figure~\ref{fig:userstudy}). Crucially, in 80\% of our generated results, participants reported a more "perfectly frozen" effect compared to the original captures. This validates that our approach successfully captures motions and eliminates artifacts to deliver on its target application.

\vspace{-1mm}
\section{Conclusion}
\label{sec:cls}
\vspace{-1mm}



We present Splannequin, a novel strategy for synthesizing high-fidelity frozen scenes from monocular Mannequin-Challenge footage. Its architecture-agnostic regularization uses temporal anchoring to identify and stabilize Gaussians, eliminating artifacts. Evaluations show Splannequin delivers superior perceptual quality with real-time rendering. This work makes high-fidelity, user-selectable freeze-time reconstruction accessible from consumer video captures.

\vspace{2pt}
\noindent {\bf Limitations.}
Our method assumes nearly static scenes and fails under rapid, non-rigid changes. Fast-moving shadows, illumination shifts, or large motions lack reliable temporal anchors, causing artifacts. Quantitative analysis of motion thresholds and frame-position dependence remains for future work, along with more adaptive anchoring strategies for challenging cases.



\newpage
\paragraph{Acknowledgements.}
This research was funded by the National Science and Technology Council, Taiwan, under Grants NSTC 112-2222-E-A49-004-MY2 and 113-2628-E-A49-023-. The authors are grateful to Google, NVIDIA, and MediaTek Inc. for their generous donations. Yu-Lun Liu acknowledges the Yushan Young Fellow Program by the MOE in Taiwan.

{
    \small
    \bibliographystyle{ieeenat_fullname}
    \bibliography{main}
}

\end{document}